\documentclass[runningheads]{llncs} %
\usepackage[T1]{fontenc}
\usepackage{graphicx}

\usepackage[caption=false]{subfig} %

\usepackage{array}
\newcolumntype{x}[1]{>{\centering\arraybackslash\hspace{0pt}}p{#1}} %
\usepackage{multirow}
\usepackage{booktabs}

\usepackage{todonotes}

\newcommand{\modelname}{Serafim}
\newcommand{\ptpt}{\mbox{PTPT}}
\newcommand{\ptbr}{\mbox{PTBR}}

\begin{document}
\title{Open Sentence Embeddings for Portuguese\\with the Serafim PT* encoders family}
\titlerunning{Open Sentence Embeddings for Portuguese}
\author{Luís Gomes \and António Branco \and João Silva \and João Rodrigues \and Rodrigo Santos}
\authorrunning{L. Gomes et al.}
\institute{%
    University of Lisbon\\
    NLX---Natural Language and Speech Group, Department of Informatics\\
    Faculdade de Ciências, Campo Grande, 1749-016 Lisboa, Portugal\\
\email{\{luis.gomes, antonio.branco, jrsilva, jarodrigues, rsdsantos\}@fc.ul.pt}}
\maketitle              %
\begin{abstract}  %
Sentence encoder encode the semantics of their input, enabling key downstream applications such as classification, clustering, or retrieval.
In this paper, we present \modelname, a family of open-source sentence encoders for Portuguese with various sizes, suited to different hardware/compute budgets.
Each model exhibits state-of-the-art performance and is made openly available under a permissive license, allowing its use for both commercial and research purposes.
Besides the sentence encoders, this paper contributes a systematic study and lessons learned concerning the selection criteria of learning objectives and parameters that support top-performing encoders.

\keywords{Sentence encoder \and Large Language Model \and Portuguese.}
\end{abstract}

\section{Introduction}
\label{sec:introduction}

Under the mainstream paradigm for Natural Language Processing (NLP), meaning is ultimately represented under the form of a vector, or embedding, for linguistic forms of any length be it a word, a multi-word sentence, a multi-sentence paragraph or a multi-paragraph text, etc.
And under the mainstream neural architecture for NLP, the Transformer~\cite{vaswani:2017:transformer}, a representation of the meaning of a linguistic input can be obtained out of the latent representations on its layers for that input,
for instance by picking the representation at a top layer.

This architecture has unfolded into different families of encoder-decoders, decoders and encoders. Despite the outstanding visibility that the decoder models have deservedly garnered, especially with the availability of ChatGPT for the general public, the models of the encoder family have not lost their traction as they have maintained a competitive performance in non-generative tasks, especially in those tasks primarily related to classification \cite{Zhong:2022:Vega}.\footnote{As a way of confirmation of this remark, the top performing model in the SuperGLUE benchmark (\url{https://super.gluebenchmark.com/leaderboard}) is an encoder, namely the Vega v2 model \cite{Zhong:2022:Vega} at the time of this writing.}

Accordingly, encoders have been the type of Transformer resorted to when the objective is to obtain a most performant representation of the meaning, aimed at being embedded in some further downstream tasks. Though this may concern any type of linguistic input---words, sentences, paragraphs, etc.---, the literature has referred to this as sentence encoding, or sometimes sentence embedding, a terminology that we will also abide by here.

Getting at the meaning representation of linguistic expressions per se is a necessary step when its is crucial to determine the degree of semantic synonymy, similarity or relatedness among them.
A prototypical use case is found at the downstream task of semantic-driven information retrieval, where at some point the input query and the relevant snippets to be retrieved are eventually screened and selected on the basis of the respective meaning representations.

With a regular encoder, the typical approach to compare sentences is to concatenate them, using a separator token, and feed that to the model.
The resulting embeddings are then provided to a separate layer that learns to classify those sentences with respect to their semantic relation.
For some tasks, however, this approach has the drawback of requiring a large number of comparisons, which easily becomes intractable.
For instance, in the prototypical information retrieval scenario, any new query would be paired with every candidate snippet to be retrieved---which in real usage scenarios have a very large volume---so that the embeddings of their concatenation can be calculated and classified.

To deal with this, standalone sentence embeddings are calculated by relying on a siamese encoder, an approach introduced by sBERT~\cite{reimers:2019:sbert} that became known in the literature as sentence encoder.
Given two sentences, these are fed, one at a time, to the same encoder, and the resulting embeddings are pooled into two embeddings and passed through a loss function designed to bring closer together in vector space the embeddings of semantically similar sentences.

An encoder trained in this way learns to map input into embeddings that live in a latent space where distance reflects semantic similarity.
This allows to semantically compare sentences, through their embeddings, using metrics such as cosine distance, thus leveraging a wide range of applications such as semantic classification, clustering, retrieval, etc.
Returning to the prototypical retrieval scenario, instead of calculating the embedding of every query-candidate pair for each new query, a sentence encoder can pre-calculate standalone embeddings for every candidate in the data set, decoupled from any particular query.

Adding to the difficulty of accessing sufficient computational resources, most NLP research focuses on English, which is just one of the around 7,000 idioms on the planet.
Consequently, there is a lack of competitive and open sentence encoders specifically developed for the vast majority of languages other than English, which happens to be also the case for Portuguese.
To the best of our knowledge, there are a couple of publicly published encoders that were developed specifically for Portuguese.
However, as we discuss in this paper, they present drawbacks,
including in what concerns their sub-optimal performance level.

Prompted by this motivation,
we report here on a systematic study concerning sentence encoding for Portuguese. As a result, we present \modelname, a family of sentence encoder models, available in different sizes, that establish a new state of the art for the respective core benchmarks of semantic text similarity (STS) and information retrieval (IR) in Portuguese.
With three versions, for 100 Million, 335M and 900M parameters, these encoders are an open code, open access and open license
collection of top performing sentence encoders in their class
specifically developed for Portuguese,
covering both the European variant, spoken in Portugal (PTPT) and the American variant, spoken in Brazil (PTBR).\footnote{Serafim models are available at \url{https://huggingface.co/PORTULAN}}

The remainder of this paper is structured as follows:
Section~\ref{sec:relatedwork} discusses related work; 
Section~\ref{sec:methodology} describes the method proposed in this study;
Section~\ref{sec:data} introduces the data sets and base models used;
Section~\ref{sec:results} reports on the experiments and discusses the results obtained;
and finally, Section~\ref{sec:conclusion} closes this paper with concluding remarks and considerations on future work.

\section{Related Work}
\label{sec:relatedwork}

This section presents work related to the goal of this paper, namely sentence encoding for Portuguese.

\textbf{Sentence encoders not specifically for Portuguese}
There exist embedders able to handle Portuguese as their underlying encoder is multilingual and had some exposure to Portuguese during its pre-training. Besides these, it is also interesting to consider encoders for the best resourced language, that is English.

To guide the choice of which of these models to compare with our own, we rely on the information provided on the SBERT site, with exhaustive information on sentence encoding,\footnote{\url{https://www.sbert.net/docs/pretrained\_models.html}} and on the results reported in the Massive Text Embedding Benchmark (MTEB) Leaderboard~\cite{muennighoff:2022:mteb}.\footnote{\url{https://huggingface.co/spaces/mteb/leaderboard}}
We selected a few that are best either of their class or under some interesting criterion.
These models and associated criterion are listed in Table~\ref{tab:chosenmultilingualmodels}, and their full names are in Table~\ref{tab:model-names}.

\begin{table}[tp]
    \centering
    \caption{Sentence encoders not specifically for Portuguese. Second column reports sizes of models (parameters) and embeddings (bits).}
    \label{tab:chosenmultilingualmodels}
    \begin{tabular}{l @{\hspace{1.5ex}} r@{/}r @{\hspace{2ex}} l}
        \toprule
        Model & \multicolumn{2}{l}{Size} & Reason for choice \\
        \midrule
        distiluse 135 multilingual & 135M &  512 & best multilingual model$^\dag$  \\ %
        mpnet 109 en               & 109M &  768 & best English model$^\dag$       \\ %
        gte 137 en                 & 137M &  768 & best under 250M$^\ddag$         \\ %
        gist embedding 109 en      & 109M &  768 & 2nd-best under 250M$^\ddag$     \\ %
        gte 434 en                 & 434M & 1024 & best under 1B$^\ddag$           \\ %
        mini 23 en                 &  23M &  384 & fast, close to best$^\dag$      \\ %
        \midrule
        \multicolumn{4}{l}{$^\dag$According to SBERT site. $^\ddag$According to MTEB benchmark.} \\
        \bottomrule
    \end{tabular}
\end{table}

\textbf{The STJ-IRIS family of encoders for Portuguese}
Aimed at supporting a semantic search system for Supremo Tribunal de Justiça (Supreme Court of Justice), the STJ-IRIS encoders~\cite{melo:2023:stjiris} are built on top of the Legal BERTimbau encoder, a BERTimbau Large~\cite{sousa:2020:bertimbau} model adapted to the legal domain.

The encoders in this STJ-IRIS family vary in terms of the methods and data used for their creation.
Two unsupervised methods are used in the adaptation to the legal domain, namely the standard masked language modeling (MLM) and TSDAE~\cite{wang:2021:tsdae}.
The sentence embedding training that follows also varies in terms of the task and training objective, resulting in over 20 models in this family.

For the sake of reference and comparison, we select the STJ-IRIS models in Table~\ref{tab:chosenstjirismodels}, whose scores on their respective HuggingFace model card indicates them as having the best performance on the various datasets used in their study.

\begin{table}[tp]
    \centering
    \caption{Chosen STJ-IRIS models.}
    \label{tab:chosenstjirismodels}
    \begin{tabular}{l @{\hspace{2ex}} l @{\hspace{2ex}} l}
        \toprule
        Model & Best on (Pearson score) & Aliased here as \\
        \midrule
        mlm-mkd-nli-sts-v1        & IRIS-STS (0.850) & stjiris 335 top irtis-sts \\
        tsdae-mkd-nli-sts-v1      & STS-B (0.855)    & stjiris 335 top sts-b \\
        mlm-gpl-nli-sts-MetaKD-v0 & ASSIN (0.811)    & stjiris 335 top assin \\
        tsdae-mkd-nli-sts-v0      & ASSIN2 (0.841)   & stjiris 335 top assin2 \\
        \bottomrule
    \end{tabular}
\end{table}

It is worth noting that these encoders were trained with data from a specific domain, and thus not all available PT data was used.
Also, recent methods for sentence encoding that are known to support better performance---viz. CoSENT~\cite{su:2022:cosent}, AnglE~\cite{li:2023:angle}, GISTEmbed~\cite{solatorio:2024:gistembed}--- were not used.
All this suggests that there is likely room for improvement. %

\textbf{Other models for Portuguese}
On HuggingFace, one can find a few other models that are presented as having been specifically trained for Portuguese, though they are not documented in some peer reviewed publication.

One finds, for instance, jmbrito/ptbr-similarity-e5-small,\footnote{\url{https://huggingface.co/jmbrito/ptbr-similarity-e5-small}} which results from fine-tuning of intfloat/multilingual-e5-small on ASSIN2.
It is reported to achieve a score of 0.79934, though unclear under what metric.
We also find a set of 22 sentence embedding models under the name MTEB-PT.\footnote{\url{https://huggingface.co/mteb-pt}}
Instead of being built upon BERT-like contextual embeddings, like the sentence encoders mentioned above, they all appear to use simple, non-contextual word embeddings (Word2Vec, Glove, etc).
No performance scores are reported.

As pointed out, these models lack a peer-reviewed publication or do not report performance scores and as such we will not consider them for our study.

\section{Methodology}
\label{sec:methodology}

The SBERT architecture is the mainstream approach to sentence encoders. It is rather straightforward and has not changed since its introduction in~\cite{reimers:2019:sbert}.
Much of the research has focused instead on the development of new training methods, in particular developing loss functions that can be used with different types of data, as each has specific use-cases, strengths and weaknesses.

\textbf{Supervised methods}
We performed two types of supervised training. The first takes as input data pairs of sentences annotated with a similarity score, and uses a loss function based on the distance between the outcome of the similarity among sentence embeddings (measured with a vector similarity function such as cosine) and the label (scaled to the range of the selected similarity function).
This is known as the Semantic Text Similarity (STS) task. 

We experimented with three selected loss functions.
The first is the canonical cosine similarity loss, proposed in the original SBERT paper, which we used as starting point and baseline for our exploratory experiments.
The other two loss functions are the recently proposed CoSENT~\cite{su:2022:cosent} loss and the AnglE~\cite{li:2023:angle} loss, which are tweaked versions of the same working principle as the first one.
Both CoSENT and AnglE perform significantly better than the original cosine similarity loss, but very close to each other in all trial experiments. 

Another type of supervised training makes use of either pairs of semantically close sentences or triples containing one anchor sentence, one positive sentence (semantically close to the anchor) and one negative sentence (either unrelated or semantically opposite to the anchor).
Unlike the previous method, here no scores are given to quantify the relatedness of input sentences.
This is the type of training supported by the Multiple Negatives Ranking (MNR) loss~\cite{henderson:2017:mnrl}, which, more formally, takes as input a set of positive pairs $(a_i, b_i)$ which is augmented with negative pairs $(a_i, b_j)$, obtained by pairing each $a_i$ with all $b_j$ sentences, assuming that any two sentences not paired at the input are unlikely to be related.  %
Without a score for quantifying the semantic closeness of positive and negative pairs, this loss function maximizes the difference between the cosine similarities of the positive and negative pair embeddings, thus bringing the embeddings of positive pairs closer together than the embeddings of negative ones.

Building on the MNR loss, GISTEmbed~\cite{solatorio:2024:gistembed} brings in a \emph{guide model}, which is another sentence encoder used to generate embeddings for every sentence and filter out negative pairs that have a similarity too close (or higher) than positive pairs.
By filtering out these negative pairs, GISTEmbed improves the signal-to-noise ratio of the training data, and thus the performance of the resulting model, at the expense of additional memory and compute time for the guide model.  %

In our exploratory experiments, we compared the MNR and GISTEmbed losses and opted for the latter because, despite requiring more time to train on the same data amount, it has shown significantly better performance on fixed-data trial experiments.  To train the final model, we used a small interim model trained with a combination of the other methods as guide. The iterim model performed comparably to the STJ-IRIS models, despite being 3 times smaller.

\textbf{Unsupervised methods}
To mitigate the scarcity of data annotated in a way that is suitable to the training of sentence encoders, we also experimented with methods that do not require labeled data.
Typically, these do not deliver results as good as supervised methods and as such we use unsupervised methods not as an alternative to supervised ones but as a supplement to them, allowing us to get some extra (pre-)training from unlabeled data.
We experimented with Transformer-based Sequential Denoising Auto-Encoder (TSDAE)~\cite{wang:2021:tsdae} and Contrastive Tension Loss with In-Batch Negatives (CT)~\cite{carlsson:2021:ct}.

TSDAE relies on the concept of a denoising auto-encoder.
Given a sentence, the approach introduces some noise into it, through the deletion of random tokens, before calculating its sentence embedding.
It then tries to reconstruct the original, denoised sentence from the embedding.

CT uses the same technique as MNR to automatically create negative pairs from positive ones.
However, instead of using positive pairs of closely related sentences, this method only requires individual sentences, which are paired with themselves.
Two slightly different instances of the model are trained simultaneously and, for each input pair, one instance is used to encode the first sentence and the other to encode the second, producing similar but not identical embeddings for identical sentences.
Both models are trained with a binary cross-entropy objective, whereby positive pairs should score 1 and negative pairs 0.
In the end, one of the models is discarded and the other is kept as the resulting model.

\textbf{Sequential transfer learning}
Each of the final Serafim models result from a combination of three training methods, applied sequentially in four stages with different types of data.  Based on our experiments, the following order produces the best performing models:

First, the pre-trained base encoder models are trained for one epoch with Contrastive Tension Loss with In-Batch Negatives, using a set of selected high-quality Portuguese corpora, already segmented at sentence level. %

Then, the models are trained for 10 epochs using GISTEmbed on data obtained by converting available NLI datasets to (anchor, positive, negative) triples --this conversion is described in the next section along with the NLI datasets.  
The validation splits of all the STS datasets are combined and used as validation set to select the best model checkpoint.

The third training stage employs the training splits of all available STS datasets combined, for 20 epochs, using both the CoSENT and AnglE loss functions separately. 
Like in the previous stage, the STS validation splits are used to choose the best performing checkpoint. 
The performance on the validation set is also used to choose the best of the two models, one trained with CoSENT and the other with AnglE.  
The models produced at this stage are the Serafim models for Semantic Textual Similarity tasks.

In the fourth and final stage, the models are trained for one epoch on a large information retrieval dataset, consisting of \emph{(query, relevant passage, irrelevant passage)} triples that are fed to the GISTEmbed training method. The resulting models are the Serafim models for Information Retrieval.

\section{Data and base models}
\label{sec:data}

\textbf{Raw text}
For the unsupervised training of our models, we made use of several of the Portuguese corpora from the OPUS Collection~\cite{tiedemann:2012:opus}. While these are parallel corpora, we take only the Portuguese part.
We made this choice as the corpora from this collection are generally well-curated and have better quality than some monolingual web crawl. We selected the following corpora:

\begin{description}
    \item[EUbookshop v2] \cite{site:eubookshop}: Corpus of documents from the Publications Office of the European Union with 4,172K sentences and 102.6M tokens, PTPT variant.
    \item[Europarl v8] \cite{koehn:2005:europarl}: Proceedings of the European Parliament with 2,000K sentences and 51.0M tokens, PTPT variant.
    \item[TED 2020 v1] \cite{reimers:2020:multilingualembeddings}: a crawl of nearly 4000 TED and TED-X transcripts from July 2020, translated by a community of volunteers, with 329K sentences and 5.2M tokens, mostly PTBR variant.
    \item[Tatoeba v2023-04-12] \cite{site:tatoeba}: database of sentences and translations, from voluntary contributions, released under CC-BY 2.0 FR, with 228K sentences and 1.6M tokens, mostly PTBR variant.
\end{description}

\textbf{Labeled data}
We train and evaluate \modelname\ on data sets, available from HuggingFace, for Semantic Text Similarity (STS), Natural Language Inference (NLI) and Information Retrieval (IR) tasks, namely:
\begin{description}
    \item[ASSIN] \cite{fonseca:2016:assin} is a data set that consists of sentence pairs manually annotated for STS (a~\mbox{1--5} score).
    Both \ptbr\ and \ptpt\ are provided, with each variant having 2.5k/0.5k/2.5k sentences for training/development/testing.
    \item[ASSIN2] \cite{real:2020:assin2} is similar to ASSIN, but it is only for \ptbr, though a larger size (6.5k/0.5k/3k) for that variant.
    \item[STSb Multi MT] \cite{site:stsbmultimt} results from the automatic translation of the STS benchmark used in the SemEval shared task~\cite{cer:2017:semeval}. Its size is 5.7k/1.5k/1.3k.
    \item[IRIS-STS] \cite{melo:2023:stjiris}, unlike the data sets above, is a very small STS data set from a specific domain (legal), that resulted from the same project that created the STJ-IRIS family of models mentioned in Section~\ref{sec:relatedwork}. This data is of the PTPT variant and its size is 1.7k/0.5k/0.5k.
    \item[PLUE] \cite{gomes:2020:plue} was obtained through automatic translation of the GLUE~\cite{wang:2018:glue} benchmark (English) to PTBR.
    It comprises several tasks, from which we use only MNLI (393k/19.6k/19.6k) and SNLI (511k/9.8k/9.8k), which are both data sets for the NLI task where pairs of sentences are labeled as \emph{entailment}, \emph{neutral} or \emph{contradiction}.
    We adapted these pairs into (anchor, positive, negative) sentence triples, for use with the GISTEmbed loss, as follows: pairs labeled as entailment (resp.\ contradiction) are considered as positive pairs (resp.\ negative), and sentences that are paired with both positive and negative sentences become anchors, resulting in a total of 534k triples.
    \item[mMARCO] \cite{bonifacio:2022:mmarco} is a multilingual version of the MS~MARCO~\cite{bajaj:2018:msmarco} data set for Information Retrieval. The translation was done automatically, using Google Translate, and includes a translation to \ptbr.
    The data set contains a collection of 8.8 million passages and 100k queries, and for each query a list of most relevant passages (qrels file).  The mMARCO training set for passage retrieval contains about 40 million of \emph{(query, relevant passage, irrelevant passage)} triples that were produced from the queries, passage collection and qrels.
    mMARCO does not comprise the MS~MARCO test set,\footnote{The test set is found in the msmarco-test2019-queries.tsv file, dowloaded from the webpage of the 2019 edition of the TREC Deep Learning Track (https://microsoft.github.io/msmarco/TREC-Deep-Learning-2019.html).} so we performed this translation ourselves, also using Google Translate.
\end{description}

\textbf{Foundation LLMs underlying \modelname ~PT*}
A sentence encoder model is built upon an existing, pre-trained encoder.
The current state-of-the-art encoders for Portuguese, which form the ``encoder ecosystem'' described in~\cite{santos:2024:ecosystem}, are the Albertina~\cite{rodrigues:2023:albertina} and the BERTimbau~\cite{sousa:2020:bertimbau} models. %
These cover a range of sizes, from 100~million to 1.5~billion parameters and, apart from BERTimbau which was pre-trained on Brazilian Portuguese (\ptbr) only, support both the \ptbr\ and European Portuguese (\ptpt) variants.

\section{Experiments and discussion}
\label{sec:results}

A selection of the main results obtained with our experiments for \textbf{Semantic textual Similarity (STS)} are in Table~\ref{tab:eval-results}.
The test data are represented in columns, with the three leftmost concerning PTPT and the remainder ones PTBR.
The encoders are represented in rows, clustered into four groups.\footnote{Their fully fledged identification is in Table \ref{tab:model-names}} 

The two groups at the top include encoders for Portuguese fine-tuned for STS: the first group with our encoders; the second group with the best representatives of previous encoders in the literature.\footnote{From the stjiris encoder collection~\cite{melo:2023:stjiris}, a particular one was selected when it was top performing among them all in at least one of the test datasets according to their Hugging Face models cards. 
It is worth noting that Pearson correlation scores are reported there, while here, Table~\ref{tab:eval-results} reports Spearman correlation scores.}

\begin{table}[h]
\centering
\caption{Results on STS, with Spearman correlation coefficient.}
\label{tab:eval-results}
\begin{tabular}{p{4cm} x{1.5cm} x{1.5cm} x{1.5cm} x{1.5cm} x{1.5cm}}
\toprule
Model                      & IRIS-STS        & STSb            & ASSIN           & ASSIN           & ASSIN2         \\
                           & ptpt            & ptpt            & ptpt            & ptbr            & ptbr           \\
\midrule
\multicolumn{6}{l}{\it Our sentence encoders} \\[.3em]

serafim 900                &         0.8231  & \textbf{0.8570} & \textbf{0.8307} &         0.7883  &         0.8267 \\
serafim 335                &         0.8226  &         0.8486  &         0.8210  & \textbf{0.7978} &         \textbf{0.8323} \\
serafim 100                & \textbf{0.8290} &         0.8350  &         0.8119  &         0.7871  &         0.8214 \\
\midrule
\multicolumn{6}{l}{\it Previous sentence encoders for Portuguese} \\[.3em]

stjiris 335 top iris-sts   &         0.7772  &         0.8486  &         0.7985  &         0.7734  &         0.8127 \\
stjiris 335 top sts-b      &         0.7873  &         0.8550  &         0.8001  &         0.7713  &         0.8160 \\
stjiris 335 top assin      &         0.7724  &         0.8364  &         0.7829  &         0.7829  &         0.8112 \\
stjiris 335 top assin2     &         0.7526  &         0.8539  &         0.7724  &         0.7724  &         0.8105 \\
\midrule
\multicolumn{6}{l}{\it Base encoders} \\[.3em]

albertina 900 ptpt         &         0.6560  &         0.5815  &         0.6183  &         0.6183  &         0.5754 \\
albertina 900 ptbr         &         0.6392  &         0.5781  &         0.6239  &         0.6239  &         0.5648 \\
bertimbau 335 ptbr         &         0.6641  &         0.5801  &         0.5445  &         0.6608  &         0.6192 \\
albertina 100 ptpt         &         0.6295  &         0.5604  &         0.6099  &         0.6099  &         0.6034 \\
albertina 100 ptbr         &         0.6279  &         0.5177  &         0.6126  &         0.6126  &         0.5875 \\
bertimbau 100 ptbr         &         0.6869  &         0.5388  &         0.5968  &         0.6442  &         0.6145 \\
\midrule
\multicolumn{6}{l}{\it Previous top-ranking multilingual and English sentence encoders} \\[.3em]

gte 434 en                 &         0.6705  &         0.6906  &         0.5147  &         0.5170  &         0.5971 \\
gte 137 en                 &         0.6977  &         0.6590  &         0.5802  &         0.5626  &         0.5503 \\
distiluse 135 multilingual &         0.7648  &         0.7687  &         0.7374  &         0.6982  &         0.7170 \\
mpnet 109 en               &         0.6484  &         0.6212  &         0.5437  &         0.5062  &         0.5631 \\
gist embedding 109 en      &         0.6736  &         0.6679  &         0.5214  &         0.5387  &         0.5450 \\
mini 23 en                 &         0.6386  &         0.6156  &         0.5664  &         0.5300  &         0.5644 \\
\bottomrule
\end{tabular}
\end{table}

The two groups at the bottom include the baseline encoders. 
The penultimate group includes the foundational encoders for Portuguese, Albertina and BERTimbau, that are the base code for our sentence encoders and that were not fine-tuned for sentence encoding. 
The last group includes state-of-the-art sentence encoders that are multilingual or specific for English.

\textbf{Portuguese datasets are crucial for having top performance} Sentence encoders specifically trained for Portuguese (top two groups) outperform sentence encoders that were not trained for Portuguese (bottom groups) but were nevertheless used to produce sentence embeddings for Portuguese. This happens by a very large margin of over some 10 to 20 points on average.

\textbf{Fine-tuning is needed for top performing sentence encoders} When fine-tuning foundational models for sentence encoding, the resulting encoders (top group) outperforms the base models (third group) by a large margin, of well over 20 points on average.

\textbf{Our sentence encoders are the state of the art for Portuguese} For every test dataset, our family of encoders (top group) outperform previous encoders for Portuguese (second group) by 0.02 (average weighted by dataset size) in terms of Spearman correlation coefficient, with the deltas: 0.0417 (IRIS-STS), 0.0020 (STSb), 0.0306 (ASSIN), 0.0149 (ASSIN), 0.0163 (ASSIN2). As expected, they also outperform both all baselines (third group) and all multilingual and English-specific alternatives (bottom group).%

\textbf{Separate training for Portuguese variants could lead to improved performance} In our exploratory experiments, we found that with the amount of data currently available for Portuguese, we got better performance by training with all data, including both language variants, than training separate models for each variant.
However, despite all Serafim models being trained with the same data, we observe that Serafim 900, based on Albertina PTPT, is better on the PTPT datasets while Serafim 335, based on BERTimbau (PTBR), takes the lead on PTBR ones. Thus, if more data becomes available in the future, training separately for each variant could lead to an increase in performance.

\textbf{Larger encoders tend to outperform smaller ones} as expected.
Our largest encoder, with 900M parameters, has the top scores in two datasets (STSb and ASSIN) and the second best scores in the others.
The second largest, 335M, always scores better than the smaller one, except for IRIS-STS, where the smaller one, 100M, actually is the top one, outperforming even the largest 900M encoder.

\textbf{Datasets offer different levels of difficulty} As expected, some datasets are easier to resolve, with 0.8570 for STSb, than others, with 0.7978 for ASSIN.
In this connection, it is worth noting that the IRIS-STS dataset is not only the smallest one but also the only one of a specific domain (legal).
All the others are of general domain.  The higher performance of \modelname\ 100, compared to the larger \modelname\ encoders, seems to indicate that the smaller model has adapted better to the specific domain, with fewer data available, than the larger ones.

\begin{table}[t]
\centering
\caption{Results on IR, with MRR@10.}
\label{tab:ir-eval-results}
\begin{tabular}{p{4cm} x{4cm}}
\toprule 
Model   & mMARCO~ptbr\\
\midrule
\multicolumn{2}{l}{\it Our sentence encoders} \\[.3em]

serafim 900 ir             & \textbf{0.8539} \\
serafim 335 ir             &         0.8278  \\
serafim 100 ir             &         0.8118  \\
\midrule
\multicolumn{2}{l}{\it Previous sentence encoders for Portuguese} \\[.3em]

stjiris 335 top assin      &         0.6197  \\
stjiris 335 top assin2     &         0.5564  \\
stjiris 335 top sts-b      &         0.5250  \\
stjiris 335 top iris-sts   &         0.4998  \\

\midrule
\multicolumn{2}{l}{\it Previous top-ranking multilingual and English sentence encoders} \\[.3em]

gte 434 en                 &         0.7414  \\
gist embedding 109 en      &         0.7227  \\
gte 137 en                 &         0.7104  \\
mini 23 en                 &         0.6875  \\
mpnet 109 en               &         0.6685  \\
distiluse 135 multilingual &         0.6494  \\
\bottomrule
\end{tabular}
\end{table}

\textbf{Information Retrieval (IR)}
A selection of the main results obtained with our experiments for IR are in Table~\ref{tab:ir-eval-results}.
The \modelname~IR models result from the fourth training stage, described in Section~\ref{sec:methodology}, using the mMARCO data (see Section~\ref{sec:data}). For evaluation, we use the MS~MARCO test set of the 2019 edition of the TREC Deep Learning Track. %
With scores clearly over 0.80, the \modelname~IR encoders outperform both the previous encoders for Portuguese and the multilingual and English encoders by a large margin, where the best one among them is the GTE encoder for English with 109M parameters, scoring 0.74.
The \modelname~encoders set thus the state of the art for Portuguese concerning IR under this dataset
and, as expected, larger models have better performance.

\section{Conclusion}
\label{sec:conclusion}

This paper presents a systematic study concerning sentence encoding for Portuguese.
Its central result is the contribution of a family of new state-of-the-art sentence encoders for this language, with 100M, 335M and 900M parameters, that are openly distributed as open source under an open license, and that serve both its European variant, spoken in Portugal, and its American variant, spoken in Brazil.
These join a wider family, which includes decoders~\cite{santos:2024:gervasio}, and data sets~\cite{osorio:2024:extraglue}. 
As future work, to further enhance encoders performance, we will seek to acquire larger volumes of data by exploring data augmentation techniques.

\begin{table}[h]
\centering
\caption{Names of the sentence encoders as they appear on HuggingFace.}
\label{tab:model-names}
\begin{tabular}{lll}
\toprule
Model                      & Name on HuggingFace                                                 & Ref.                       \\
\midrule
stjiris 335 top iris-sts   & stjiris/bert-large-portuguese-cased-legal-mlm-mkd-nli-sts-v1        & \cite{melo:2023:stjiris}        \\
stjiris 335 top sts-b      & stjiris/bert-large-portuguese-cased-legal-tsdae-mkd-nli-sts-v1      & \cite{melo:2023:stjiris}        \\
stjiris 335 top assin      & stjiris/bert-large-portuguese-cased-legal-mlm-gpl-nli-sts-MetaKD-v0 & \cite{melo:2023:stjiris}        \\
stjiris 335 top assin2     & stjiris/bert-large-portuguese-cased-legal-tsdae-mkd-nli-sts-v0      & \cite{melo:2023:stjiris}        \\
albertina 100 ptpt         & PORTULAN/albertina-100m-portuguese-ptpt-encoder                     & \cite{rodrigues:2023:albertina} \\
albertina 100 ptbr         & PORTULAN/albertina-100m-portuguese-ptbr-encoder                     & \cite{rodrigues:2023:albertina} \\
bertimbau 100 ptbr         & neuralmind/bert-base-portuguese-cased                               & \cite{sousa:2020:bertimbau}     \\
bertimbau 335 ptbr         & neuralmind/bert-large-portuguese-cased                              & \cite{sousa:2020:bertimbau}     \\
albertina 900 ptpt         & PORTULAN/albertina-900m-portuguese-ptpt-encoder                     & \cite{rodrigues:2023:albertina} \\
albertina 900 ptbr         & PORTULAN/albertina-900m-portuguese-ptbr-encoder                     & \cite{rodrigues:2023:albertina} \\
distiluse 135 multilingual & distiluse-base-multilingual-cased-v1                                & \cite{reimers:2019:sbert}       \\
mpnet 109 en               & all-mpnet-base-v2                                                   & n/a                             \\
gist embedding 109 en      & avsolatorio/GIST-Embedding-v0                                       & \cite{solatorio:2024:gistembed} \\
gte 137 en                 & Alibaba-NLP/gte-base-en-v1.5                                        & \cite{li:2023:gte}              \\
gte 434 en                 & Alibaba-NLP/gte-large-en-v1.5                                       & \cite{li:2023:gte}              \\
mini 23 en                 & all-MiniLM-L6-v2                                                    & n/a                             \\
\bottomrule
\end{tabular}
\end{table}

\begin{credits}

\subsubsection{\ackname}
This work was partially supported by:
PORTULAN CLARIN---Research Infrastructure for the Science and Technology of Language, funded by Lisboa 2020, Alentejo 2020 and FCT (PINFRA/22117/2016);
ACCELERAT.AI---Multilingual Intelligent Contact Centers, funded by IAPMEI (C625734525-00462629);
and IMPROMPT---Image Alteration with Language Prompts, funded by FCT (CPCA-IAC/AV/590897/2023).

\subsubsection{\discintname}
The authors have no competing interests to declare. %

\end{credits}

\bibliographystyle{splncs04}
\bibliography{main}

\end{document}